\documentclass[pdflatex,sn-mathphys-num]{sn-jnl}

\usepackage{graphicx}%
\usepackage{titlesec}%
\usepackage{multirow}%
\usepackage{amsmath,amssymb,amsfonts}%
\usepackage{amsthm}%
\usepackage{mathrsfs}%
\usepackage[title]{appendix}%
\usepackage{xcolor}%
\usepackage{textcomp}%
\usepackage{manyfoot}%
\usepackage{booktabs}%
\usepackage{algorithm}%
\usepackage{algorithmic}%
\usepackage{listings}%
\usepackage{adjustbox}
\usepackage{subcaption}
\usepackage{array}
\usepackage{wrapfig}
\usepackage{pifont}
\usepackage[T1]{fontenc}
\usepackage{bookmark}
\usepackage{lmodern}
\usepackage[inline]{enumitem}

\titleformat{\paragraph}[runin]%
  {\normalfont\normalsize\bfseries}%
  {\theparagraph}%
  {1em}%
  {}         

\newcommand{\cmark}{\ding{51}}%
\newcommand{\xmark}{\ding{55}}%
 
\DeclareMathOperator{\tr}{tr}
\newcolumntype{R}[2]{%
    >{\adjustbox{angle=#1,lap=\width-(#2)}\bgroup}%
    l%
    <{\egroup}%
}
\newcommand*\rot{\multicolumn{1}{R{45}{1em}}}
\newcounter{mainfig} 
\newcounter{maintab}

\begin{document}
\title[Article Title]{Uncover and Unlearn Nuisances: Agnostic Fully Test-Time Adaptation}

\author*[1]{\fnm{Ponhvoan} \sur{Srey}}\email{ponhvoan002@e.ntu.edu.sg}
\equalcont{These authors contributed equally to this work.}

\author*[2,3]{\fnm{Yaxin} \sur{Shi}}\email{shi\_yaxin@cfar.a-star.edu.sg}
\equalcont{These authors contributed equally to this work.}

\author[2,3]{\fnm{Hangwei} \sur{Qian}}\email{qian\_hangwei@cfar.a-star.edu.sg}

\author[2,3]{\fnm{Jing} \sur{Li}}\email{kyle.jingli@gmail.com}

\author[1,2,3]{\fnm{Ivor W.} \sur{Tsang}}\email{ivor.tsang@gmail.com}

\affil[1]{\orgdiv{College of Computing and Data Science}, \orgname{Nanyang Technological University (NTU)}, \orgaddress{\postcode{639798}, \city{Singapore}, \country{Singapore}}}
\affil[2]{\orgdiv{Centre for Frontier AI Research}, \orgname{Agency for Science, Technology and Research (A*STAR)}, \orgaddress{\postcode{138632}, \city{Singapore}, \country{Singapore}}} 
\affil[3]{\orgdiv{Institute of High Performance Computing}, \orgname{Agency for Science, Technology and Research (A*STAR)}, \orgaddress{\postcode{138632}, \city{Singapore}, \country{Singapore}}}
    
\abstract{
Fully Test-Time Adaptation (FTTA) addresses domain shifts without access to source data and training protocols of the pre-trained models. Traditional strategies that align source and target feature distributions are infeasible in FTTA due to the absence of training data and unpredictable target domains. In this work, we exploit a dual perspective on FTTA, and propose Agnostic FTTA (AFTTA) as a novel formulation that enables the usage of off-the-shelf domain transformations during test-time to enable direct generalization to unforeseeable target data. To address this, we develop an uncover-and-unlearn approach. First, we uncover potential unwanted shifts between source and target domains by simulating them through predefined mappings and consider them as nuisances. 
Then, during test-time prediction, the model is enforced to unlearn these nuisances by regularizing the consequent shifts in latent representations and label predictions.
Specifically, a mutual information-based criterion is devised and applied to guide nuisances unlearning in the feature space and encourage confident and consistent prediction in label space.
Our proposed approach explicitly addresses agnostic domain shifts, enabling superior model generalization under FTTA constraints. Extensive experiments on various tasks, involving corruption and style shifts, demonstrate that our method consistently outperforms existing approaches.
}
\maketitle

\section{Introduction}

Deep neural networks (DNNs) have shown excellent performance across numerous applications when training and test data are independently and identically distributed~\citep{lecun2015deep, silver2016mastering, faster2015towards}.
However, under domain shifts, where the training (source) data differ from the test (target) data, DNN
models suffer from a precipitous drop in accuracy \citep{hendrycks2019robustness}. 
Unlike the human visual system which is robust to irrelevant shifts in images, DNNs struggle with classification when the test data undergo even simple corruptions or general style changes, such as rotation, blur \citep{ovadia2019can,hendrycks2019robustness}, or synthetic-to-real \citep{peng2017visda} shifts. 

To address domain shift challenges in real-world applications, domain adaptation (DA) techniques have been widely explored. Traditional DA methods minimize the discrepancy between source and target feature distributions, promoting domain-invariant representations to improve target generalizability~\citep{ben2006analysis, song2018improving, saito2018maximum, tang2020discriminative, saito2020universal}. 
However, these approaches rely on full access to labeled source and target data, which is often infeasible due to privacy or efficiency constraints.
To overcome these limitations, more practical DA approaches have emerged. 
Test-time training (TTT)~\citep{sun2020test, liu2021ttt++} reduces the need for labeled data by incorporating auxiliary self-supervised tasks to bridge the source and target domain, but requires retraining the source model with modified training protocols, \textit{i.e.,} combining the supervised task objectives with self-supervised losses~\citep{banerjee2021self, darestani2022test}. 
Source-free domain adaptation (SFDA) eliminates the need for source data altogether \citep{li2020model, goyal2022test} by leveraging target feature structures \citep{yang2021generalized} or introducing generative modules to enforce feature invariance \citep{yeh2021sofa, kurmi2021domain}, making DA more practical and efficient.

In this paper, we target the challenging and practical setting of Fully Test-Time Adaptation (FTTA), where both the source training data and training protocol of the source model are inaccessible, and the target domain remains unforeseeable.
This scenario raises a critical question: \textit{without source data and the original training loss cannot be modified}, how can we effectively capture the distributional shifts and enable the model with improved generalization capacity for fully test-time adaptation?

To overcome this challenge, we present a dual perspective on FTTA.
We argue that, for any FTTA task, %
the agnostic transformation from source to target data can be dealt with a lazy but smart process over target data. 
That means, instead of directly working on that agnostic transformation, we can expect the adapted model to be capable of handling the potential shifts by encouraging it to be robust against any \emph{arbitrary} transformation on the accessible test data.
This insight leads to Agnostic FTTA (AFTTA), a novel reformulation of FTTA that leverages off-the-shelf domain-shift priors to enable direct generalization to unknown target domains at test-time.
Within AFTTA, we can explicitly exploit agnostic shifts between source and target domains to address fully test-time adaption tasks effectively.
The key assumption of this approach is as follows: \textit{if a model could demonstrate robustness to unknown source-target domain shifts, it should also be robust to predefined, simple off-the-shelf transformations that partially mimic these unknown shifts when applied to the target domain}.

\begin{table}[t]
\centering
\caption{Comparison of Domain Adaptation Problem Settings.
Our proposed Agnostic FTTA setting emphasizes leveraging source-target domain shifts to tackle the challenges inherent in standard FTTA. By enabling the applicability of source-target alignment strategies, it bridges the gap between FTTA and traditional approaches such as DA, TTT, and SFDA. 
Here, $x^s, y^s, x^t, y^t$  represent source data, source labels, target data, and target labels, respectively.}
\renewcommand{\arraystretch}{1.1}
\setlength{\tabcolsep}{0.9mm}
\begin{adjustbox}{width=\textwidth}
    \begin{tabular}{l|cccccc|c}
        \hline
        Problem Setting & Source data & Target Data & Training Loss & Testing Loss & Offline & Online & \multicolumn{1}{c}{\begin{tabular}[c]{@{}c@{}}Domain shift \\prior\end{tabular}} \\ \hline
        Domain Adaption (DA)~\citep{tang2020discriminative, saito2020universal}  &$x^{s}, y^{s}$  &  $x^{t}$    &   {\small$L(x^{s},y^{s}) + L(x^{t},y^{t})$} &  --  &  --  &  -- &  \cmark \\
        Test-time training (TTT)~\citep{sun2020test, liu2021ttt++}         &$x^{s}, y^{s}$  &  $x^{t}$    &   {\small$L(x^{s},y^{s}) + L(x^{t})$}            &  $L(x^{t})$           &  \cmark & \cmark  & -- \\
        Source-free DA (SFDA)~\citep{li2020model, goyal2022test}             &  \xmark        &  $x^{t}$    &   {\small$L(x^{s}_{*},y^{s}_{*}) + L(x^{t})$}      &  -- & \cmark & \xmark    &  \cmark  \\
        Fully test-time adaptation (FTTA)~\citep{wang2020tent, zhang2022memo} &  \xmark        &  $x^{t}$    &   \xmark   &    {\small$L(x^{t})$}  & \cmark & \cmark  & \xmark \\ \hline
        Agnostic FTTA (ours)                      &  \xmark        &  $x^{t}$    &   \xmark   &    {\small$L(x^{t})$}  & \cmark & \cmark  &  \cmark \\ \hline     
    \end{tabular}
\end{adjustbox}\label{tab:com_settings}
\end{table}

\begin{wrapfigure}{R}{0.45\textwidth}
  \begin{center}
    \includegraphics[width=0.44\textwidth]{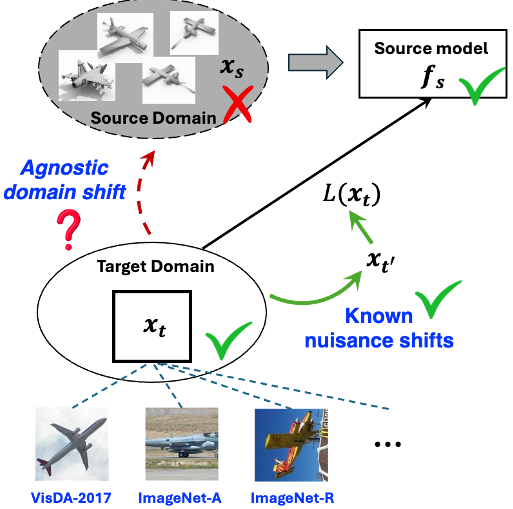}
  \end{center}
  \vspace{1mm}
  \caption{Illustration of Agnostic Fully Test-Time Adaptation (AFTTA).}
    \label{fig:tirnu_AFTTA}
\end{wrapfigure}

To address the problems of AFTTA, we adopt an uncover-and-unlearn strategy.
As the agnostic domain shift between the source and target is entirely unknown, we \textit{\textbf{uncover}} this hidden shift by simulating plausible shifts with a set of accessible off-the-shelf mapping functions (see Fig.~\ref{fig:tirnu_AFTTA}). 
These transformations are applied directly to the target data, generating perturbed samples that approximate potential latent shifts relative to the unobservable source domain.
Our motivation stems from the observation that a vanilla disruption serves as a simple yet effective choice for representing the unwanted shift, which is a nuisance factor \textit{w.r.t} the main classification task. This conjecture is theoretically grounded in our variational interpretation presented in Section~\ref{sec:uncover_and_unlearn_for_AFTTA}. 
The adapted model is then encouraged to \textbf{\textit{unlearn}} the influence of these nuisances via a loss using only the target data $L(x_t)$, promoting feature invariance during test-time inference to enhance model generalization.

To facilitate nuisance factor unlearning during adaptation, we enforce the source model to remove feature shifts in target data caused by nuisances.
Specifically, we minimize the mutual information between the learned features and the shift induced by the nuisances.
A matrix-based Renyi's entropy functional algorithm is presented to estimate that term efficiently. %
This criterion disentangles and removes the influence of nuisances during test-time feature learning, enabling the adapted model to learn shift-invariant test features that enhance generalization to the target domain, without relying on source data or training protocols.
Additionally, we regularize the model to produce confident and consistent label predictions on both the original test data and nuisance-shifted data, further reinforcing invariance of the extracted test features. 
Collectively, our proposed adaptation method, \textbf{T}est-time \textbf{I}nvariant \textbf{R}epresentation learning through\textbf{ N}uisances \textbf{U}nlearning (TIRNU), ensures shift-invariant and compact test data representations that are independent of nuisances factors that mimic source-target shifts, thereby improving model generalization to the target domain.

Our TIRNU fosters shift-invariant representations within the constraints of FTTA.
In the experiments, we demonstrate that TIRNU performs competitively with other strong baselines under FTTA. TIRNU consistently achieves state-of-the-art results on the CIFAR10-C/100-C/10.1, VisDA-2017, Imagenet-A, Imagenet-R, and Imagenet-C datasets. Furthermore, TIRNU learns better features compared to MEMO, whose objective of encouraging confident and invariant predictions is similar to ours.
Our work also delivers the key message that mimicking nuisance-augmented data that represents unwanted shifts between the source and target domains can help address practical domain shift challenges when source data is inaccessible.
Accordingly, we open up the flexibility to benefit source-free domain adaptation with arbitrary prior knowledge on the source-target domain shifts, namely, to provide shift-specific nuisance augmentations under TIRNU, enhancing the performance of the adapted model with to-the-point domain invariant features.

\section{Related Work}

\paragraph{Fully Test-Time Adaptation}, namely FTTA, is a special and challenging Domain Adaptation (DA) setting, where no source data, no source model training protocol and no test data annotation are available for model adaption at test time.
In FTTA, a key focus is defining proper objectives on the test data to enable adaptation.
Test entropy minimization (Tent) \citep{wang2020tent} facilitates FTTA using an entropy regularization, enforcing the source model to achieve confident label prediction on test data samples with either online or offline updates.
Source Hypothesis Transfer (SHOT) \citep{liang2020we} explores the offline FTTA.
It addresses source-target domain shifts with pseudo-test data labelling. A clustering objective that is strongly aligned with the main classification task is incorporated at test time for adaptation.
Recently, there has been emerging interests in exploring online FTTA \citep{zhang2023adanpc,boudiaf2022parameter,zhang2022memo}. 
For instance, ETA~\citep{niu2022efficient} achieves efficient online FTTA through sample-efficient entropy minimization.
ENT~\citep{goyal2022test} proposes conjugate pseudo-labels for online adaptation. 
SAR~\citep{niu2023towards} explores stabilizing online TTA in wild test settings.
LAME~\citep{boudiaf2022parameter} focuses on parameter-free online TTA, adjusting classifier output probabilities rather than feature extractor parameters.

Despite the wide attention on fully test-time adaptation issues, there have been few efforts specifically aimed at fostering feature-invariance under the FTTA constraints. Most of the prior approaches, especially those online methods, operate under the assumption that the difference between the source and target domains primarily lies with the first and second-order moments of the features \citep{schneider2020improving}. Therefore, they employed
test data normalization with test statistics, \textit{e.g.,} batch normalization (BN) to perform weak domain alignment, ensuring feature invariance \citep{saito2020universal,chang2019domain, zajkac2019split}. While SHOT \citep{liang2020we} addresses domain shifts via general source-hypothesis-transfer, %
its pseudo-labelling process can propagate errors that misguide the adaptation.

Our work differs in its explicit goal of uncovering and unlearning task-irrelevant variations—providing a principled route to domain-invariant representation learning under FTTA, without relying on source data or target labels. While domain-invariant feature learning has been widely explored in standard DA and SFDA settings~\citep{saito2018maximum, saito2020universal, xu2021open, li2020model, yeh2021sofa, kurmi2021domain}, these settings allow for either source supervision or target annotation. More recent efforts in multi-source domain adaptation (MSDA) and semi-supervised domain adaptation (SSDA) have proposed co-training, distillation, and multi-view learning strategies~\citep{10332198, 9868789, 9828482, DBLP:conf/iccv/NgoCKPC23, DBLP:conf/bmvc/KimNPKLC22}, but they still operate under relaxed assumptions, such as access to labeled source domains or partial target supervision. In contrast, our method is tailored for the fully test-time regime, offering a general framework for both online and offline adaptation by disentangling and suppressing nuisance variation.

\begin{figure}
    \centering
    \includegraphics[width=\textwidth]{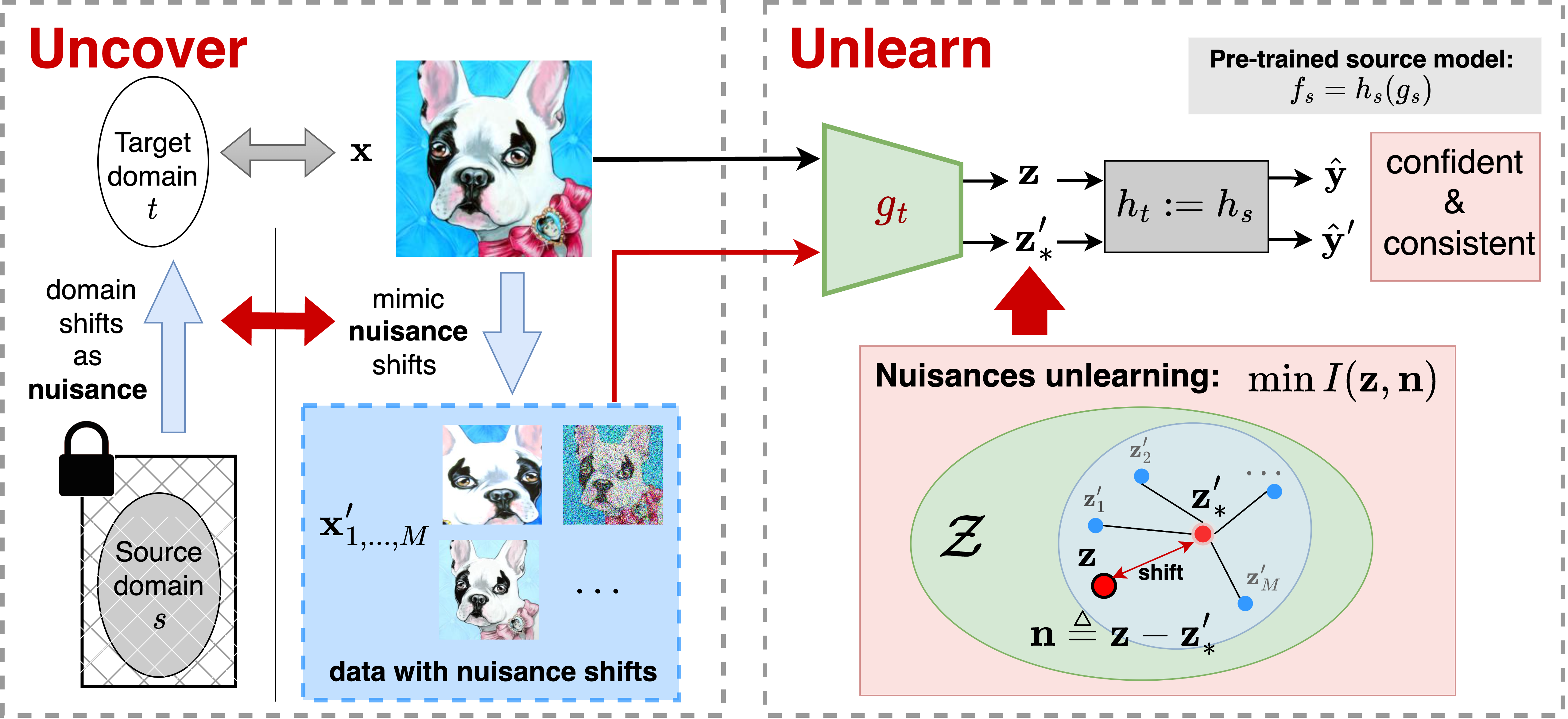}
    \caption{
    Our TIRNU approach for agnostic fully test-time domain adaptation.
    Without access to the (source) training data, we adopt an uncover-and-unlearn strategy. We uncover potential unwanted shifts between the source and target as nuisances and mimic these nuisances onto the target data with off-the-shelf data transformations. 
    During test-time adaptation, the feature extractor is updated to unlearn these nuisance shifts by mitigating their impact on feature representations, ensuring confident and consistent label predictions.
    Our model enhances the source model’s generalization to the agnostic target domain by promoting shift-invariant feature representations.
    }
    \label{fig:motivation_trinu}
    \vspace{-4mm}
\end{figure}

The most relevant approach to ours is MEMO \citep{zhang2022memo}.
MEMO explores a \textit{test-time robustification} issue, namely to improve model robustness without accessing the model training procedure.
It focuses on one-shot online updates and ensures model prediction invariance across multiple augmentations of a test sample.
Our work, depicted in Fig.~\ref{fig:motivation_trinu}, differs from MEMO in two key aspects. 
First, instead of targeting model robustness, we emphasize the insight to enhance test-time adaptability by identifying and eliminating task-uninformative nuisance shifts.
Second, rather than paying mere attention to prediction invariance, TIRNU prioritizes shift-invariant feature extraction through latent nuisance unlearning. This offers a novel criterion for improved test-time adaptation performance.

\paragraph{Nuisance Factors Disentanglement.} %
{A nuisance factor is any random variable that affects the observed data but is not informative to the task of interest \citep{achille2018emergence}. 
In a classification task, random data corruptions or style changes that commonly trigger dataset shifts through data poisoning~\citep{jia2021intrinsic,steinhardt2017certified} or translation~\citep{wang2021rethinking,shu2021encoding}, introduce nuisances $n$ that shifts the input image $x$ yet is uninformative for predicting the label $y$. %
Prior works have particularly leveraged the property of nuisances to benefit their objective tasks with invariant features. 
Theoretically, Alemi et al. proposed a feature learning algorithm that empirically shows improved resistance to adversarial nuisances~\citep{alemi2016deep}.
Achille et al. explored feature invariance to nuisances of the Variational Information bottleneck under a task-nuisance decomposition assumption~\citep{achille2018emergence}.
Application-wise, Yao et al. and Zheng et al. explored nuisance disentanglement in deep clustering to learn feature partitioning that disregards specific nuisance factors~\citep{yao2023sanitized, zeng2023deep}. %
Similarly, Wei et al. improved model robustness with nuisance-label supervision on top of the main task ~\citep{wei2021nuisance}. These methods require the nuisance factor to be predefined in an explicit form of labels or attributes, which brings in extra annotation burdens.}
{In our work, we eliminate the explicit definition of nuisance factors while directly exploiting the influence of nuisances instead. We uncover unwanted shifts performed with augmentations as nuisances and collectively unlearn them by removing their triggered feature shift and label inconsistency.}

\section{Uncover and Unlearn Nuisances for Agnostic Fully Test-time Adaptation}

Consider a source distribution $P_\mathcal{S} = p(x_s, y_s)$ and a target distribution $P_\mathcal{T}=p(x_t, y_t)$, both defined over the joint space $\mathcal{X} \times \mathcal{Y}$, where $\mathcal{X}$ represents the input space and $\mathcal{Y}$ the label space. 
Data-label pairs $(x_s, y_s)$ and $(x_t, y_t) $ are sampled from domain $\mathcal{S}$ and $\mathcal{T}$, respectively.
In the context of fully test-time adaptation, a source-trained model 
$f_s: \mathcal{X}_{s} \to \mathcal{Y}_{s}$ is initially learned from $\mathcal{S}$, while at test time, only unlabelled data $x_t$ from the unforeseeable test domain $\mathcal{T}$ is available for adaptation and inference.

The core challenge in all domain adaptation problems lies in addressing distribution shifts, where $p(x_s, y_s) \ne p(x_t, y_t)$.
This issue becomes even more complex in fully test-time adaptation, where both the source training data and training protocol of the source model are inaccessible.
Facing the absence of source data and unforeseeable test data, bridging the discrepancy between source and target distributions becomes significantly harder, yet it remains essential for guiding the source model to adapt effectively and deliver accurate, reliable predictions on the target distribution.

\subsection{Agnostic Fully Test-time Adaptation}

To overcome the aforementioned challenge, we adopt the covariate shift assumption and present a dual perspective on FTTA by particularly considering the agnostic shift between the source and target distributions.

Specifically, we first assume that the joint distribution discrepancy occurs only in the input space $\mathcal{X}$, while the conditional distributions remain the same. That is, 
\begin{equation}
   p(x_{s}) \ne p(x_{t}), \; p(y| x_{s}) = p(y|x_{t}). 
\end{equation}
\noindent {Under FTTA, where the source distribution is unavailable, we propose to ground the distribution alignment strategies from DA by forming a duality perspective on FTTA.}

\paragraph{Proposition:}
For any Fully Test-Time Adaptation (FTTA) task where a model {\small$f_s: \mathcal{X}_{s} \to \mathcal{Y}_{s}$} trained on source domain $\mathcal{S}$ is tested on a unforeseen target domain \mbox{$\mathcal{T} = \{(x_t, y_t) | x_t \in \mathcal{X}_{t}, y_t \in \mathcal{Y}\}$}, there exists a domain-agnostic transformation \mbox{$g: \mathcal{X}_{t} \to \mathcal{X}_{s}$}, such that
\begin{itemize}[itemsep=1mm]
    \item $g$ maps the target data distribution $P_{\mathcal{T}}$ to some points or a subspace in the source distribution $P_{\mathcal{S}}$ under a well-defined distance metric $d$, \textit{i.e.} $d(P_{\mathcal{S}},P^{g}_{\mathcal{T}}) < \epsilon$, where  $P^{g}_\mathcal{T}$ is the distribution of $g(\mathcal{X}_{t})$ and $\epsilon$ is a small constant.
    
    \item the composition $f\circ g: \mathcal{X}_{t} \to \mathcal{Y}$ could achieve comparable performance as \mbox{$f:\mathcal{X}_{s} \to \mathcal{Y}$}, preserving the predictive capacity of $f$ on the target domain.

    \item $g$ is domain-agnostic in the sense that it can be learned or approximated with only the feed test data for adaptation, without explicit supervision on $\mathcal{Y}_t$.
    
\end{itemize}

\vspace{2mm}
Based on that proposition, we present Agnostic FTTA (AFTTA) as a reformulation of standard FTTA that specifically explore the domain-agnostic transformations to facilitate direct generalization to the unknown target domain at test-time with available domain-shift priors (refer to Table~\ref{tab:com_settings}). Suppose the potential distribution shift between $p(x_{t})$ and $p(x_{s})$ is induced by an agnostic transformation denoted as $T_{t\to s}$. 
The corresponding density ratio for this transformation is defined as
\begin{equation}
    r_{t\to s} = \frac{p(x_{s})}{p(x_{t})}
\end{equation}

\subsection{Uncover and Unlearn Nuisances for Agnostic FTTA}\label{sec:uncover_and_unlearn_for_AFTTA}

To deal with the agnostic shift in the absence of source domain data, we propose to 
\begin{enumerate*}
\item[1).] \textit{\textbf{Uncover}:} approximating the agnostic shift between $\mathcal{S}$ and $\mathcal{T}$ using a predefined set of off-the-shelf mapping functions that serves as domain-shift priors. These mappings are then applied to the accessible test data to simulate the nuisance shift onto the target domain.
\item[2).] \textit{\textbf{Unlearn}:}  During test-time adaptation, we focus on isolating and mitigating the influence of these shifts on the target data, ensuring effective adaptation of the pre-trained model from the mimicked source-target shift.
\end{enumerate*}

We apply a set of off-the-shelf mapping functions to construct a transformation that potentially partially mimics the agonistic source-to-target shift.
This transformation is applied to $p(x_{t})$, producing a shifted target domain $\mathcal{T}^{'}$ with distribution $p(x_{t^{'}})$ where
\begin{equation}
\label{eq:transform}
    T_{t\to t^{'}} = T_{k} \circ T_{k-1} \circ \dots \circ T_{1}
\end{equation}
represents the composition of multiple pre-defined simple transformations.
The density ratio corresponding to this transformation is given by:
\begin{equation}
    r_{t\to t^{'}} = \frac{p(x_{t^{'}})}{p(x_{t})}
\end{equation}

This explicit shift-modelling approach enables the simulation and analysis of potential transformations between source and target distributions, supporting effective adaptation of the pre-trained model to account for these distributional changes.
In this \textbf{\textit{uncover}} context, the density ratio $r_{t\to t^{'}}$ quantifies the extent to which the simplified transformation $T_{t\to t^{'}}$ modifies the target distribution, serving as a partial approximation of the full complex source-target shift represented by
$r_{t\to s}$.

For the \textbf{\textit{unlearn}} phase at test-time, suppose we technically remove the influence of $T_{t\to t^{'}}$ during test-time inference, the adopted unlearning mechanism directly adjusts for ${r_{t\to t^{'}}}$ while recovering the test-domain distribution $p(x_{t})$, \textit{i.e.},
$p(x_{t})= {p(x_{t^{'}})}/{r_{t\to t^{'}}}$.
In this case, the test-time adjustment caused by influence removal of $T_{t\to t^{'}}$
enables $f_s$ to operate directly on $p(x_{t})$, while the remaining challenge is reduced to addressing the residual source-target transformation:
\begin{equation}
    r_{\text{residual}} = r_{t\to s}/ r_{t\to t^{'}}\label{eq:ratio_res}
\end{equation}
The residual ratio isolates the portion of $T_{t\to s}$ not captured by $T_{t\to t^{'}}$, thereby simplifying the complexity gap between the source and target distribution, making the overall adaptation easier even without accessing the source training data.

Theoretically, suppose the source model $f_s$ is trained to minimize the expected loss under the source distribution $p(x_{s})$, defined as
\begin{equation}
    \mathbb{E}_{p(x_{s})} \; [\mathcal{L}(f_s)]
\end{equation}
where $\mathcal{L}$ indicates the loss function associated with $f_s$.
The source-target transferability is defined as the difference between $f_s$'s expected error on the source domain and its generalization on the target domain, that is,
\begin{equation}
    \varepsilon (f_s; p(x_{s}),p(x_{t})) = \mathbb{E}_{p(x_{t})} [\mathcal{L}(f_s)] - \mathbb{E}_{p(x_{s})} [\mathcal{L}(f_s)].
\end{equation}
When the simpler predefined transformation $T_{t\to t^{'}}$ is applied to $p(x_{t})$, the density ratio ${r_{t^{'} \to s}}$ represents the discrepancy between $p(x_{s})$ and $p(x_{t^{'}})$. The generalization error of $f_s$ on $p(x_{t^{'}})$ becomes
\begin{equation}
    \varepsilon (f_s; p(x_{s}),p(x_{t^{'}})) = \mathbb{E}_{p(x_{t^{'}})} [\mathcal{L}(f_s)] - \mathbb{E}_{p(x_{s})} [\mathcal{L}(f_s)].
\end{equation}
Since $T_{t\to t^{'}}$ simplifies part of $T_{t\to s}$,
$r_{t\to t^{'}}$ reduces the complexity of ${r_{t^{'}\to s}}$, making the generalization error smaller than directly adapting from $p(x_{s})$ and $p(x_{t})$. That is

\begin{equation}
    \varepsilon (f_s; p(x_{s}),p(x_{t})) = \varepsilon (f_s; p(x_{s}),p(x_{t^{'}})) + \Delta
\end{equation}
where $\Delta$ is the residual error due to $T_{t\to s}/T_{t\to t^{'}}$. 
The closer $T_{t\to t^{'}}$ approximates $T_{t\to s}$, the greater the transferability $f_s$ achieves to $p(x_{t})$ through nuisances-unlearning during test-time adaptation.

In summary, applying a simpler off-the-shelf transformation $T_{t\to t^{'}}$ onto $p(x_{t})$ helps decompose the complex agnostic source-target shift $T_{t\to s}$.
By removing the nuisances effect of $T_{t\to t^{'}}$ in source-model adaptation at test-time, the residual transformation $T_{t\to s}/T_{t\to t^{'}}$ is isolated. This simplification reduces the complexity of adaptation from $p(x_{s})$ to $p(x_{t})$, making the process manageable without accessing the source data.


\paragraph{\textbf{Remark: A Variational Perspective on Uncover-and-Unlearn for AFTTA.}}

Our uncover-and-unlearn strategy can be interpreted through a variational lens. 
Under the Agnostic FTTA setting, where both the source distribution $p(x_s)$ and the domain shift $T_{t \to s}$ are unknown, we model the domain shift as a latent nuisance transformation variable $T$, which maps test samples to a hypothetical source domain where the model $f_s$ was trained~\cite{xie2020unsupervised}.
Suppose the true distribution over such transformations is $p(T)$, the predictive distribution for a target sample $x_t$ is thus marginalized over the unknown transformation distribution:
\[
p(y \mid x_t) = \int p(y \mid T(x_t)) \, p(T) \, dT.
\]
Introducing a tractable surrogate distribution $q(T)$ and applying Jensen’s inequality~\cite{kingma2013auto} yields a variational lower bound:
\begin{equation}\label{eq:variational}
\log p(y \mid x_t) \geq \mathbb{E}_{T \sim q(T)} \left[ \log p_\theta(y \mid T(x_t)) \right] - \text{KL}(q(T) \| p(T)).
\end{equation}
This provides a principled explanation for our strategy: defining $q(T)$ over a set of off-the-shelf transformations $T_{t \to t'}$ serves as a prior/basis for the unknown shift distribution~\cite{xie2020unsupervised}. 
The choice of $T \sim q(T)$ at test time is therefore not arbitrary or heuristic; rather, it is grounded in a principled variational approximation of the true prediction marginalized over the nuisance variable.
Moreover, since $T$ governs the shift between domains, the divergence $\text{KL}(q(T) \| p(T))$ implicitly reflects distributional mismatch, connecting to the residual ratio $r_{\text{residual}}$ in Eq.~\eqref{eq:ratio_res}.

This formulation encourages a model to marginalize out nuisance factors that may obscure task-relevant signals without requiring source data access. 
While enforcing consistency across $x_T$ and its transformed variants $t(x_T)$ at the \textit{output level} alone may not fully eliminate nuisance factors~\cite{goyal2022test}, we emphasize \textit{feature-level invariance} by explicitly unlearning nuisance-sensitive components in the representation space to reinforce this variational approximation for enhanced model generalization and adaptation performance.
This dual mechanism —\textbf{uncovering} plausible shifts through $q(t)$ and \textbf{unlearning} their influence —forms the foundation of our model design in the absence of source supervision, test-domain priors, or source model configurations.

\section{Method}
\label{sec:method}

{Based on the aforementioned principles, we propose to address the agnostic source-target transformation in full test-time adaptation with some predefined process over target data.
That means, instead of working directly on that agnostic transformation $T_{t\to s}$, we expect the adapted model to be capable of handling the potential shifts by encouraging it to be against any
arbitrary transformation $T_{t\to t^{'}}$ on accessible test data.}

In this section, we begin with an overview of the problem statement and the design of our method. 
Subsequently, we present TIRNU as a simple but effective implementation of our uncover-and-unlearn nuisance insight by 
\begin{enumerate*} 
    \item[1)]using a set of common mapping functions for image augmentations, $\mathcal{A}$,
    to represent the transformation $T_{t\to t^{'}}$ in Eqn~\ref{eq:transform} to \textit{uncover} the source-target shift, and
    \item[2)] introducing a mutual information-based objective to \textit{unlearn} nuisances induced by the test-time augmentations.
\end{enumerate*} Finally, we present the optimization algorithm, which leverages a matrix-based Renyi's entropy functional for mutual information estimation. 

\subsection{Problem Statement}

In fully test-time adaptation, we aim to learn an adapted target model {\small$f_t: \mathcal{X}_t \to \mathcal{Y}_t$} from the source model $f_s$, such that it can accurately infer $y_t$ from $x_t$.
Without loss of generality, we assume the DNN-parameterized source model
$f_s$ consists of two modules: the feature extractor $g_s: \mathcal{X}_s \to \mathcal{Z}_s$ where $z \in \mathcal{Z}_s \subseteq \mathbb{R}^{d}$ denotes the latent feature of sample $x$ and the classification module $h_s: \mathcal{Z}_s \to \mathcal{Y}_s$ where $y \in \mathcal{Y}_s \subseteq \mathbb{R}^K$ indicates the data label ($d$ and $K$ indicate the feature dimension and number of classes, respectively). Therefore, $f_s(x) = h_s (g_s(x))$. 
Similarly, for the adapted model on the target domain, $f_t(x) = h_t(g_t(x))$. 
During test-time adaptation, we freeze the classification module, \textit{i.e.,} $h_t=h_s$, and seek to learn an adapted feature extractor $g_t^{\theta}$, parameterized by $\theta$, that can learn shift-invariant features for the target test data without accessing the source data, and without modifying the training protocol of $f_s$.

\subsection{Uncover Nuisances}
To address the problem, we explore the agnostic source-target shift with predefined simple mapping functions and treat them as nuisances that are uninformative to the main task of the source model. 
During test-time adaptation, the model should be updated to be invariant to the influence of these nuisances to seek good classification performance. 
Therefore, we uncover those nuisances at the test time with test data transformations forming test data set with nuisances shifts.

To be specific, under the FTTA setting, where no prior knowledge of the source-target shift is given,
we collectively mimic undesired shifts between the source and target domains through widely adopted self-supervision methods, including data augmentation techniques like random cropping or transformation \citep{chen2020simple}, $L_p$ data corruptions \citep{siedel2023investigating}, and AugMix \citep{hendrycks2019augmix}.
These test-data augmentations involve nuisance factors that would derange the source model from learning informative features for the target domain data but should not trigger differences in the final label prediction for the given test samples. %
This property enables us to perform source-free invariant feature learning by involving those nuisance shifts in the test data and subsequently unlearning the influences of nuisances during the test-time feature extraction phase.

Suppose that these augmentation strategies define a collective set of mapping functions, which is denoted as $\mathcal{A}$, conveying unwanted domain or data shifts. This augmentation set $\mathcal{A}$ corresponds to the transformation $T_{t\to t^{'}}$ that partially mimics the source-target shift.
Given a test sample $x$, we uniformly sample $M$ transformations $a \sim \mathcal{U}(\mathcal{A})$ and obtain $M$ nuisance-shifted versions of $x$, denoted as $\{x'_{j}\}_{j=1}^{M}$, where $x'_{j}= a_{j} (x)$. 
We explicitly define the influence of the nuisances shift on $x$ as the collective feature shifts caused by those unwanted data transformations, which is denoted as
\begin{equation}
\begin{split}
    n & \triangleq \mathbb{E}_{\mathcal{U}(\mathcal{A})}\left[ g^{\theta}_{t}(x) - g^{\theta}_{t}\left(a(x)\right) \right] \\
               & \approx g^{\theta}_{t}(x) - \frac{1}{M} \sum_{j=1}^{M} g^{\theta}_{t}\left(x'_{j}\right) = z - z'_{*},
\end{split}
\end{equation}

\noindent where $z'_{*} = \frac{1}{M} \sum_{j=1}^{M} g^{\theta}_{t}\left(x'_{j}\right)$ is the centroid of the shifted features $z'_{j} = g^{\theta}_{t}\left(x'_{j}\right)$ for $j=1, \ldots, M$.
This new variable $n$ captures the overall feature shift caused by the nuisances.  
The adapted model should unlearn these negative influences to achieve shift-invariant test features for model generalization.

\subsection{Unlearn Nuisances}

To unlearn the influence of nuisances at test time, we propose to minimize a nuisance unlearning loss $\mathcal{L}_{NU}$ at the feature space which regularizes the mutual information between the extracted test data features and the collective feature shift, \textit{i.e.},
\begin{equation}
\label{eq:min_latent_nuisance_unlearn}
    \mathcal{L}_{NU} = I(g^{\theta}_{t}(x);n). %
\end{equation}
This criterion forces the model to disentangle and remove the influence of nuisances when extracting informative test features for domain invariance.
Through this loss, the model is adapted to be robust to the specified shift defined in $\mathcal{A}$. 
Under mild assumptions, by minimizing $\mathcal{L}_{\text{NU}}$, the representation $z=g^{\theta}_{t}(x)$ becomes more informative of the main task, and invariant to nuisances (refer to Appendix \ref{sec:obj}).

Moreover, to enhance the shift-invariance of the extracted test features, we further encourage the adapted model to maintain confident and consistent label predictions on the nuisance-shifted test data. 
To be specific, let $\hat{y}$ denote the label prediction of $x$ with $f_{t}$, \textit{i.e.}, $\hat{y} = h_{t}(z)$. 
Similarly, for the nuisance-shifted data $x'$, we have $\hat{y}' = h_{t}(z')$. 
We employ the vanilla entropy regularizer to form the constraint for confident prediction on both the vanilla and nuisance-shifted data.
We further constrain the model to maintain consistent label predictions on the original test feature and the nuisance-shifted features with cross-entropy loss terms~\cite {goyal2022test}. The individual influence of label confidence and label consistency loss terms are examined in model ablation presented 
in Section \ref{sec:analysis}.
Collectively, our additional regularizer for invariant label prediction
\begin{equation}
\label{eq:min_label}
    \mathcal{L}_{label} = \sum_{z,z'} H(\hat{y}) + H(\hat{y}') + CE (\hat{y}, \hat{y}') + CE (\hat{y}', \hat{y}),
\end{equation}
where {\small$H(\hat{y}) = -\sum_{c} p(\hat{y}_c) \log p(\hat{y}_c)$}, {\small$H(\hat{y}') = - \sum_{c} p(\hat{y}'_c) \log p(\hat{y}'_c)$}, 
{\small$CE(\hat{y}, \hat{y}') = -\sum_{c} p(\hat{y}_c) \log p(\hat{y}'_c)$} and {\small$CE(\hat{y}', \hat{y}) = -\sum_{c} p(\hat{y}'_c) \log p(\hat{y}_c)$}.
The last term can be omitted in scenarios where transformations are deemed too strong to be reliable~\cite{sohn2020fixmatch}.
For all terms, %
$p$ denotes the predicted probability. 

Therefore, the overall objective of our TIRNU approach is 
\begin{equation}
\label{eq:overall_objective}
    \min_{\theta} \; 
    \mathcal{L}_{NU} + \lambda \mathcal{L}_{label} 
\end{equation}
where $\lambda >0$ is a hyperparameter balancing the terms of explicit nuisance unlearning in the feature space and the regularization of confident and consistent label prediction. 

\paragraph{Remark}: TIRNU presents a general framework to solve  FTTA via uncovering and unlearning nuisances (refer to Section~\ref{sec:uncover_and_unlearn_for_AFTTA}).
Within the framework, functionally, we highlight the incorporation of %
$\mathcal{L}_{NU}$ as a general criterion to benefit model robustness of test-time adaptation approaches. %
Application-wise, we suggest arbitrarily crafted domain-shift prior set $\mathcal{A}$ to involve possible hints of source-target shifts in practical scenarios.
We present experiments with different configurations of $\mathcal{A}$ in Section~\ref{subsec:augmentation}.

The objective of TIRNU serves as a tractable surrogate to the variational formulation in Eq.~\ref{eq:variational}, supporting our Uncover-and-Unlearn strategy. Under FTTA, where neither source data nor labels are available, the expected log-likelihood is approximated via prediction consistency across augmentations, enforced by $\mathcal{L}_\text{label}$.
For the KL regularization term, we consider the domain shift as composed of task-relevant semantics ($s$) and nuisance factors ($n$). Accordingly, we interpret the divergence term $\mathrm{KL}(q(T)\parallel p(T))$ as reflecting mismatches in both semantic and nuisance distributions. Since $s$ is unobserved in FTTA, we focus on suppressing the nuisance-induced discrepancy. Therefore, minimizing the nuisance-unlearning loss $\mathcal{L}_\text{NU} = I(z; n)$ acts as a partial realization of the KL regularization, targeting the portion of feature variation entangled with nuisance transformations.
As detailed in Appendix~\ref{sec:obj}, this promotes $I(z; s)$ under mild assumptions, enhancing semantic fidelity without supervision. Unlike conventional alignment losses, our formulation explicitly encourages disentanglement, enabling robust generalization under agnostic, multimodal shifts.

\subsection{Optimization}

The optimization of our TIRNU objective is initially challenging since the proposed nuisance unlearning loss $\mathcal{L}_{NU} = I(g^{\theta}_{t}(x);n)$ in Eq.(2) is computationally intractable.
To tackle the issue, we leverage matrix-based Rényi's $\alpha$-order entropy functionals \citep{renyi1961measures} to estimate this mutual-information-based loss term, allowing a direct and efficient optimization of our proposed approach.

The matrix-based entropy functionals provide a useful means of estimating variable dependencies without explicit estimation of the underlying data distributions, while in a differentiable manner \citep{yu2021measuring}. Unlike mutual information estimators that rely on variational approximation or adversarial training, such as MINE~\citep{belghazi2018mutual}, a matrix-based estimator offers a clear mathematical definition and computational efficiency, particularly for larger networks~\citep{zhang2022multi}.
With a differentiable matrix-based mutual information estimator for the $\mathcal{L}_{NU}$ term in our objective, we can directly update TIRNU using a gradient-based optimizer.

To be specific, our nuisance unlearning criterion $\mathcal{L}_{NU} = I(g^{\theta}_{t}(x);n) = I(z;n)$ can be rewritten as
\begin{equation}
\label{eq:info}
    I(z;n) = H(z) + H(n) - H(z, n) 
\end{equation}
\begin{algorithm}[t]
\caption{Test-time Invariant Representation learning through Nuisances Unlearning (TIRNU)}
\label{algo:tirnu}
\begin{algorithmic}[1]
\REQUIRE Source pre-trained model $f_s = h_s \circ g_s$, test data $\mathcal{X}_t = \{(x_t^{i}) \}_{i=1}^{N_t}$, augmentation $\mathcal{A}$.
\WHILE{Adaptation} 
    \STATE $\mathcal{X} \leftarrow$ SampleMiniBatch$(\mathcal{X}_t)$ \\
    \STATE Get augmented inputs $\mathcal{X}' \leftarrow$ a$(\mathcal{X})$, where $a \sim \mathcal{U} (\mathcal{A})$ \\
    \STATE $\mathcal{Z}, \mathcal{Z}' \leftarrow$ Forward $(\mathcal{X}, \mathcal{X}', g_s)$\\
    \STATE $\mathcal{Y}, \mathcal{Y}' \leftarrow$ Forward $(\mathcal{Z}, \mathcal{Z}', h_s)$\\
    \STATE Estimate $\mathcal{L}_{NU}$ with Equations \ref{eq:info}, \ref{eq:ent}, \ref{eq:jent}, and compute $\mathcal{L}_{label}$ with Equation \ref{eq:min_label}\\
    \STATE Update $g_s$ via SGD with objective $\mathcal{L}_{NU} + \lambda \mathcal{L}_{label} $
\ENDWHILE
\end{algorithmic}
\end{algorithm}
Inspired by ~\citep{zhang2022multi}, we estimate each entropy term $H$ using the matrix-based Rényi's $\alpha$-order entropy functionals, $H_{\alpha}$, which is precisely Shannon's entropy when $\alpha \rightarrow 1$. 
Given a mini-batch of $B$ data samples, we obtain both random vectors $z$ and $n$ from the adapted feature extractor $g^{\theta}_{t}$.
According to~\citep{giraldo2014measures}, the entropy of $z$ can be defined over the eigenspectrum of a normalized Gram matrix $A_{z} = K_{z}/\tr({K_{z}}) \in \mathbb{R}^{B \times B} $, where $K_{z}(i,j)=\kappa(z_i,z_j)$ and $\kappa$ is the Gaussian kernel. That is,
\begin{equation}
\label{eq:ent}
    H_{\alpha}(A_{z}) = \frac{1}{1-\alpha}\log_2 \tr{(A_{z}^{\alpha})} = \frac{1}{1-\alpha}\log_2 \left(\sum_{j=1}^{n} \lambda_j(A_{z})^{\alpha} \right),
\end{equation}
\noindent where $\lambda_j$ is the $j$-th eigenvalue of $A_{z}$. The entropy of $n$ can be similarly defined based on another normalized Gram matrix \mbox{$A_{n} \in \mathbb{R}^{B \times B}$}. Further, the joint entropy for $z$ and $n$ can be defined as
\begin{equation}
\label{eq:jent}
    H_{\alpha}(z,n) = H_{\alpha}\left(\frac{A_{z} \circ A_{n}}{\tr(A_{z} \circ A_{n})}\right),
\end{equation}
\noindent where $\circ$ denotes the Hadamard product. 
With Eq.\eqref{eq:info}-\eqref{eq:jent}, 
the matrix-based Rényi's $\alpha$-order mutual information estimation $I_{\alpha}(z;n)$ in analogy of Shannon’s mutual information is given by:
\begin{equation}
\label{eq:info_alpha}
    I_{\alpha}(z;n) = H_{\alpha}(A_{z}) + H_{\alpha}(A_{n}) - H_{\alpha}(A_{z}, A_{n}) 
\end{equation}
which we estimate $I(z;n)$ with $I_\alpha$ by setting a value of $\alpha$ close to 1.
We empirically testified that our model performance is not sensitive to this $\alpha$ hyperparameter (see Fig.~\ref{fig:meib_par}). Algorithm \ref{algo:tirnu} presents the overall scheme of our TIRNU approach for fully test-time adaptation with a batch update.

\section{Experiments}

\paragraph{Datasets} 
To assess our method's robustness against real-world image distortions across diverse domains, we conduct evaluations in three key settings: mixed image corruptions, natural adversarial examples, and style transfer scenarios.
These three scenarios cover a wide range of domain shifts studied in the literature. 
Our experiments are conducted on seven different distribution shift benchmark datasets: 
CIFAR10-C/CIFAR100-C \citep{hendrycks2019robustness}, CIFAR10.1 \citep{recht2019imagenet}, Imagenet-A \citep{hendrycks2021natural}, Imagenet-R \citep{hendrycks2021many}, Imagenet-C \citep{hendrycks2019robustness}, and VisDA-2017 \citep{peng2017visda}.

\paragraph{Baselines}
Our TIRNU allows both offline and online updates for adaptation. 
Here, we compare our approach to prior test-time adaptation methods under both problem settings.
We compare with TestBN, SHOT, TENT, and TT-C under the offline source-free adaption setting; 
and compare with MEMO, SAR, and CPL under the online setting for model updates with streaming data.

\begin{itemize}
    \item \textbf{Source}: the pre-trained source model that is evaluated on the test data without any adaptation.
    
    \item Test-time batch normalization (\textbf{Test BN}) \citep{schneider2020improving}: estimate the test data statistics and use it to normalize the BN layers to update the source model.
    
    \item Source hypothesis transfer (\textbf{SHOT}) \citep{liang2020we}: on top of test entropy, minimizes a global diversity term and a pseudo-labelling term, updating the whole source feature extractor.
    
    \item Test entropy minimization (\textbf{Tent}) \citep{wang2020tent}: Tent minimizes the test entropy, and updates only the affine parameters of the BN layers in the source model. %

    \item Test-time contrastive learning (\textbf{TT-C}) \citep{liu2021ttt++}: an FTTA ablated version of TTT++. While TTT++ requires source data information, we implement TT-C to merely minimise an auxiliary contrastive loss%
    , updating the source feature extractor and the self-supervised head.

    \item Marginal entropy minimization with one test point { (\textbf{MEMO}) 
    \citep{zhang2022memo}}: an online FTTA method that performs one-shot updates and adapts models using entropy from marginal output distribution over augmentations of the test data.
    
    \item Sharpness-aware and reliable entropy minimization method (\textbf{SAR})~\citep{niu2023towards}: an online FTTA method that performs stabilized updates with a tailor-designed optimization scheme.

    \item Conjugate pseudo-labels (\textbf{CPL})~\citep{goyal2022test}: proposed an unsupervised loss with pseudo-labels based on the convex conjugate formulation for online test-time adaptation.
    
\end{itemize}
Implementation details are provided in Appendix \ref{sec:implementation}.

\subsection{Image Corruptions in a Mix}

To investigate robustness to corruptions, we experiment on the CIFAR10-C and CIFAR100-C datasets, which contain 15 different types of algorithmic corruptions, such as the noise and blur effects, applied on the CIFAR datasets. 
Given a source model pre-trained on the clean CIFAR10/CIFAR100 data, 
in addition to the normal evaluation of the test error on the dataset with each of the corruption types, we further constructed two more challenging datasets that comprise a mixture of 5 and 10 different corruptions, respectively.
The results are summarized in Table \ref{tab:corrupt}. Note that, for the test sets with only a single corruption each, we report the average error across all corruption types for each method.

Table \ref{tab:corrupt} shows that TIRNU consistently outperforms its competitors by a considerable margin. For example, across the CIFAR10-C dataset, TIRNU achieves 13\%, 11\%, and 2\% improvements on the average, mixture of 5, mixture of 10 corruptions respectively, over the best baselines. For the CIFAR100-C dataset, TIRNU similarly shows competitive accuracy, although for the mixture of 10 corruptions, all methods are unable to produce satisfactory results.

\begin{table*}[ht]
\caption{Summary of error (\%) on the CIFAR10-C/CIFAR100-C datasets. The average error across all corruption types, and the error on the mixed datasets with 5 and 10 corruptions, are reported. We report the mean error and standard deviation for TIRNU over 5 runs.}
\label{tab:corrupt}
    \begin{adjustbox}{width=\textwidth,center}
        \begin{tabular}{{p{0.14\textwidth}>{\centering}p{0.14\textwidth}>{\centering}p{0.14\textwidth}>{\centering}p{0.14\textwidth}>{\centering}p{0.14\textwidth}>{\centering}p{0.14\textwidth}>{\centering\arraybackslash}p{0.14\textwidth}}}
             \toprule
             \multirow{2}{*}{\textbf{Method}} & \multicolumn{3}{c}{\textbf{CIFAR10-C}} & \multicolumn{3}{c}{\textbf{CIFAR100-C}} \\
             \cmidrule(lr){2-4}
             \cmidrule(lr){5-7}
             & \textbf{Average} & \textbf{Mix 5} & \textbf{Mix 10} & \textbf{Average} & \textbf{Mix 5} & \textbf{Mix 10} \\
             \midrule
             \textbf{Source} & 33.27 & 44.85 & 35.30 & 59.37 & 69.27 & 98.95 \\ 
             \textbf{Test BN} & 17.35 & 29.90 & 29.35 & 43.41 & 57.74 & 99.00 \\
             \textbf{SHOT} & 15.21 & 25.77 & 26.59 & 38.24 & 52.69 & \textbf{98.84} \\ 
             \textbf{Tent} & 16.06 & 29.77 & 33.18 & 37.77 & 55.53 & 99.42 \\
             \textbf{TT-C} & 15.27 & 28.64 & 30.46 & 39.68 & 58.40 & 98.90\\
             \midrule
             \textbf{TIRNU} & \mbox{\textbf{13.13} \tiny{($\pm$ 0.32)}} & \mbox{\textbf{23.68} \tiny{($\pm$ 0.18)}} & \mbox{\textbf{25.69} \tiny{($\pm$ 0.32)}} & \mbox{\textbf{36.68} \tiny{($\pm$ 0.08)}} & \mbox{\textbf{51.62} \tiny{($\pm$ 0.04)}} & \mbox{99.08 \tiny{($\pm$ 0.10)}} \\
             \bottomrule
        \end{tabular}
    \end{adjustbox}
\end{table*}

\subsection{Natural and Adversarial Examples}

For adversarial examples, we have the Imagenet-A as representatives for the Imagenet datasets. 
We additionally test on CIFAR10.1 for natural temporal shift in the CIFAR dataset. Meanwhile, Imagenet-R covers an extensive range of realistic and challenging domain shifts, such as sculptures, sketches, tattoos, toys, and origami.

In the CIFAR10.1 experiment shown in Table \ref{tab:natural}, the baselines are unable to successfully adapt the model the test data as all their errors increase 6-21\% compared to the source model. Meanwhile, TIRNU is the only method to improve over the source model by 3\%. For Imagenet-A, while the dataset is difficult for all methods, ours presents a clear advantage over the other baselines, being the only one with an error falling below 99\% (Table \ref{tab:natural}). 
The experiments on Imagenet-R indicate similar positive results, with TIRNU beating the baselines (see Table \ref{tab:natural}).

\begin{table}[!htb]
  \caption{Error (\%) on CIFAR10.1, Imagenet-A, and Imagenet-R}
  \label{tab:natural}
  \centering
    \begin{tabular}{l c c c}
         \toprule
         \textbf{Method} & \textbf{CIFAR10.1} & \textbf{Imagenet-A} & \textbf{Imagenet-R}\\
         \midrule
         \textbf{Source} & 14.00 & 99.97 & 63.84 \\
         \textbf{Test BN} & 17.00 & 99.32 & 67.83 \\
         \textbf{SHOT} & 16.60 & 99.44 & 57.11 \\ 
         \textbf{Tent} & 15.90 & 99.37 & 61.73 \\
         \textbf{TT-C} & 14.80 & 99.43 & 61.58 \\
         \midrule
         \textbf{TIRNU} & \textbf{13.65} & \textbf{98.59} & \textbf{56.20}\\
         \bottomrule
    \end{tabular}
\end{table}

\subsection{Style Transfer}

We consider the VisDA-2017 and Imagenet-R datasets for the style transfer task. 
The experiments on the synthetic-to-real VisDA-2017 are presented in Table \ref{tab:visda}. We find that TIRNU performs the most consistently, reaching the lowest error on five classes, and second lowest on four. Overall, TIRNU ranks first in terms of accuracy, and the per-class error improves by 3\% over the best baseline. 

\begin{table*}[ht]
\caption{Error (\%) on VisDA-2017 dataset. The best result (lowest error) is bolded while the second best is underlined. The methods are ranked, and the average rank is given in the last column.}
\label{tab:visda}
    \begin{adjustbox}{width=\textwidth,center}
        \begin{tabular}{l c c c c c c c c c c c c |c c}
             \toprule
             \textbf{Method} & \textbf{plane} & \textbf{bcycl} & \textbf{bus} & \textbf{car} & \textbf{horse} & \textbf{knife} & \textbf{mcycl} & \textbf{person} & \textbf{plant} & \textbf{skate} & \textbf{train} & \textbf{truck} & \textbf{Per Class} & \textbf{Rank}\\
             \cmidrule(lr){1-13}
             \cmidrule(lr){14-15}
             \textbf{Source} & 48.44 & 88.58 & 67.72 & \textbf{33.25} & 75.95 & 97.20 & 18.79 & 96.60 & 49.31 & 85.40 & \textbf{21.13} & 99.86 & 58.92 & 4.83 \\
             \textbf{Test BN} & 57.38 & 57.15 & 34.35 & 58.60 & 44.28 & 54.12 & 19.67 & 65.85 & 47.07 & 64.62 & 36.71 & \textbf{85.87} & 51.80 & 4.33 \\
             \textbf{SHOT} & \underline{16.81} & \textbf{37.53} & 31.00 & 46.40 & 20.38 & 41.88 & 15.42 & 32.68 & 18.09 & \textbf{51.73} & \underline{24.06} & 96.03 & 37.14 & 2.33 \\
             \textbf{Tent} & 17.77 & 73.61 & \textbf{23.97} & 45.36 & 26.20 & \textbf{22.75} & 11.02 & 45.92 & 28.75 & 76.55 & 34.89 & 99.69 & 42.05 & 3.17\\
             \textbf{TT-C} & 73.07 & \underline{39.51} & 46.20 & 56.25 & 35.17 & 69.54 & 20.84 & 54.15 & 25.50 & 56.95 & 50.92 & 87.58 & 50.55 & 4.25\\
             \midrule
             \textbf{TIRNU (Ours)} & \textbf{15.41} & 60.00 & \underline{27.36} & \underline{38.72} & \textbf{16.20} & \underline{34.12} & \textbf{10.59} & \textbf{31.12} & \textbf{17.94} & \underline{56.12} & 24.58 & 99.82 & \textbf{36.03} & \textbf{2.08} \\
             \bottomrule
        \end{tabular}
    \end{adjustbox}
\end{table*}

\begin{table}[h]
\centering
    
\end{table}
\newpage
\subsection{Online Adaptation} 

\begin{wrapfigure}{R}{0.5\textwidth}
  \begin{center}
    \includegraphics[width=0.48\textwidth]{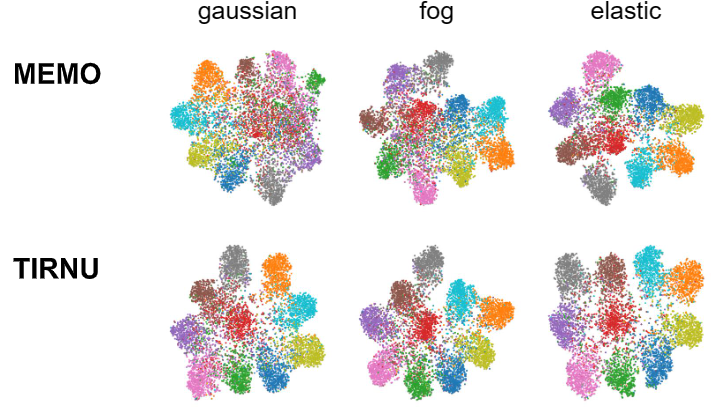}
  \end{center}
  \caption{t-SNE visualization of features extracted by MEMO and TIRNU on CIFAR10-C.}
    \label{fig:tirnu_v_memo}
\end{wrapfigure}

We empirically show that TIRNU is competitive with recent online adaptation baselines. We compare to MEMO \citep{zhang2022memo}, SAR \citep{niu2023towards}, and CPL \citep{goyal2022test} on the CIFAR10-C and Imagenet-C datasets. The error (\%) on the highest corruption level of CIFAR10-C and Imagenet-C are shown in Table \ref{tab:online}. Our experiments indicate that TIRNU is capable of online adaptation, as it outperforms CPL and MEMO, and achieves competitive results with SAR. As Fig. \ref{fig:tirnu_v_memo} shows, TIRNU learns considerably more compact representation than MEMO, leading to lower error. In particular, MEMO's objective of achieving confident and invariant prediction aligns with ours. This highlights TIRNU's advantage in constraining the model to unlearn nuisances in the feature space.

\vspace{-1em}
\begin{table}[!hbt] 
\caption{Online adaptation error (\%) on the CIFAR10-C and Imagenet-C datasets.}
\label{tab:online}
    \begin{adjustbox}{width=\textwidth,center}
        \begin{tabular}{l c c c c c c c c c c c c c c c c}
              & \multicolumn{16}{c}{\textbf{CIFAR10-C}} \\  \toprule
             \textbf{Method}& \rot{gaussian} & \rot{shot} & \rot{impulse} & \rot{defocus} & \rot{glass} & \rot{motion} & \rot{zoom} & \rot{snow} & \rot{frost} & \rot{fog} & \rot{brightness} & \rot{contrast} & \rot{elastic} & \rot{pixel} & \rot{jpeg}& \rot{Avg} \\ \midrule
             
            \textbf{Source} & 64.26          & 57.71          & 80.27          & 11.34         & 33.78          & 17.18          & 9.58          & 24.65          & 27.79          & 36.99          & 13.19         & 53.17          & 17.26          & 30.89          & 21.06          & 33.27          \\
            \textbf{SAR} & 23.09          & 20.78          & 32.67          & 10.48         & 21.75          & 13.46          & 8.65          & 16.85          & 15.42          & 20.85          & 10.81         & 15.64          & 16.86          & 11.21          & 17.88          & 17.09          \\
            \textbf{MEMO} & 41.28          & 37.80          & 57.70          & 11.02         & 27.23          & 15.28          & 9.27          & 18.82          & 19.12          & 29.04          & 11.14         & 37.82          & 16.15          & 18.83          & 18.64          & 24.61          \\
            \textbf{CPL} & 22.83          & 20.53          & 32.09          & 10.35         & 21.88          & 13.02          & 8.68          & 16.87          & 15.06          & 20.12          & 10.57         & 15.22          & 16.80          & 11.04          & 17.34          & 16.83          \\ \midrule
             \textbf{TIRNU} & \textbf{19.69} & \textbf{18.20} & \textbf{28.96} & \textbf{9.86} & \textbf{20.63} & \textbf{12.58} & \textbf{8.42} & \textbf{15.54} & \textbf{13.59} & \textbf{17.75} & \textbf{9.71} & \textbf{13.22} & \textbf{15.25} & \textbf{10.35} & \textbf{16.41} & \textbf{15.34} \\ \bottomrule
        \end{tabular}
    \end{adjustbox}
    
    \begin{adjustbox}{width=\textwidth,center}
        \begin{tabular}{l c c c c c c c c c c c c c c c c}
             
              & \multicolumn{16}{c}{\textbf{Imagenet-C}} \\  \toprule
             \textbf{Method}& \rot{gaussian} & \rot{shot} & \rot{impulse} & \rot{defocus} & \rot{glass} & \rot{motion} & \rot{zoom} & \rot{snow} & \rot{frost} & \rot{fog} & \rot{brightness} & \rot{contrast} & \rot{elastic} & \rot{pixel} & \rot{jpeg}& \rot{Avg} \\ \midrule
             
            \textbf{Source}    & 97.79          & 97.07          & 98.15          & 82.08          & 90.18          & 85.22          & 77.51          & 83.12          & 76.69          & 75.57          & 41.06          & 94.57          & 83.05          & 79.38          & 68.35          & 81.99 \\
\textbf{SAR}       & \textbf{72.13}          & \textbf{71.93}          & \textbf{71.11}          & \textbf{73.22}          & \textbf{73.80}          & 60.99          & 52.29          & 55.04          & \textbf{59.20}          & 43.72          & 32.69          & \textbf{65.56}          & 47.05          & 42.86          & 49.15          & \textbf{58.05} \\
\textbf{MEMO}      & 91.90          & 90.88          & 89.73          & 78.74          & 85.79          & 80.33          & 72.17          & 75.89          & 70.60          & 63.84          & 37.87          & 89.53          & 72.35          & 65.21          & 59.68          & 74.97 \\
\textbf{CPL} & 82.11          & 81.10          & 81.52          & 82.25          & 82.13          & 69.37          & 56.88          & 61.40          & 63.63          & 47.34          & 33.32          & 78.45          & 51.85          & 46.30          & 54.71          & 64.82 \\ \midrule
\textbf{TIRNU}     & 73.90 & 74.13 & 73.04 & 73.43 & 78.83 & \textbf{59.20} & \textbf{50.99} & \textbf{51.97} & 61.43 & \textbf{42.67} & \textbf{31.80} & 74.53 & \textbf{45.95} & \textbf{41.72} & \textbf{48.04} & 58.78 \\ \bottomrule
        \end{tabular}
    \end{adjustbox}
\end{table}
\vspace{-1em}

\subsection{Ablation Study}
\label{sec:analysis}

Here, we explore the individual effects of our proposed nuisance unlearning loss $\mathcal{L}_{NU}$, and the regularizer $\mathcal{L}_{label}$. 
To this end, we evaluate three ablated versions of TIRNU on CIFAR10-C:
\begin{enumerate}
    \item TIRNU without (w/o) $\mathcal{L}_{NU}$, \textit{i.e.} includes only $\mathcal{L}_{label}$;
    \item TIRNU w/o $\mathcal{L}_{label}$, \textit{i.e.} includes only $\mathcal{L}_{NU}$;
    \item TIRNU w/o $H(\hat{y}'), CE(\hat{y},\hat{y}'), CE(\hat{y}',\hat{y})$, \textit{i.e.}, includes our $\mathcal{L}_{NU}$ and  the $\lambda H(\hat{y})$ term adopted in \citep{wang2020tent, liang2020we};
\end{enumerate}
    The results are reported in Table \ref{tab:ablate}. As a reference, the errors (\%) of the best baseline, SHOT, and the full TIRNU model, are indicated. 

\begin{table*}[!hbt] 
\caption{Error(\%) on CIFAR10-C by different ablated versions of TIRNU. The final TIRNU model with the overall objective yields the best results consistently against all corruption types.}
\label{tab:ablate}
    \begin{adjustbox}{width=\textwidth,center}
        \begin{tabular}{l c c c c c c c c c c c c c c c c}
             \toprule
             \textbf{Approach} & \rot{snow} & \rot{gaussian} & \rot{motion} & \rot{elastic} & \rot{bright} & \rot{pixelate} & \rot{defocus} & \rot{shot} & \rot{impulse} & \rot{glass} & \rot{zoom} & \rot{frost} & \rot{fog} & \rot{contrast} & \rot{jpeg}& \rot{Avg} \\
             \midrule
            
             \textbf{SHOT} & 19.91          & 17.90          & 28.90          & 9.82          & 20.20          & 12.33          & 8.09          & 15.35          & 13.89          & 17.46         & 9.74         & 12.83          & 15.60          & 10.42         & 16.19          & 15.24          \\
            \textbf{w/o $\mathcal{L}_{NU}$}  & 19.78          & 17.96          & 28.90          & 9.88          & 20.00          & 12.36          & 8.00          & 15.38          & 13.76          & 17.27         & 9.82         & 12.86          & 15.37          & 10.35         & 16.11          & 15.19          \\
            \textbf{w/o $\mathcal{L}_{label}$} & 20.38          & 18.37          & 31.54          & 9.44          & 20.72          & 11.48          & 8.17          & 14.48          & 12.99          & 16.06         & 8.85         & 12.65          & 14.98          & 10.14         & 16.77          & 15.13          \\
            \textbf{w/o $H(\hat{y}'), CE(\hat{y},\hat{y}'), CE(\hat{y}',\hat{y})$} & 19.33          & 17.43          & 27.89          & 9.3           & 19.19          & 11.48          & 7.75          & 13.86          & 12.98          & 15.71         & 9.08         & 12.33          & 14.48          & 10.15         & 15.56          & 14.43      \\ \midrule
             \textbf{TIRNU (Full)} & \textbf{16.61} & \textbf{15.32} & \textbf{24.53} & \textbf{8.95} & \textbf{18.17} & \textbf{10.58} & \textbf{7.48} & \textbf{12.85} & \textbf{12.09} & \textbf{13.3} & \textbf{8.1} & \textbf{10.59} & \textbf{14.21} & \textbf{9.85} & \textbf{14.39} & \textbf{13.13}  \\ \bottomrule
        \end{tabular}
    \end{adjustbox}
\end{table*}

From Table \ref{tab:ablate}, we observe that our proposed information loss $\mathcal{L}_{NU}$ performs competitively with SHOT, and on average, achieves slightly lower classification error on the CIFAR10-C corruptions. 
Likewise, when only the label regularizer $\mathcal{L}_{label}$ is used, we obtain comparable performance to SHOT. 
When we combine $\mathcal{L}_{NU}$ with $H(\hat{y})$, the error drops significantly to 14.43\%, indicating the importance of the information term.
In the full model, we see again another improvement of 8\%, outlining the contribution of the extra terms $H(\hat{y}'), CE(\hat{y},\hat{y}'), CE(\hat{y}',\hat{y})$ in the regularizer.

\subsection{Effects of Augmentation}
\label{subsec:augmentation}

As the augmentation $\mathcal{A}$ is a crucial component in our approach, to understand its effectiveness, we experiment with various off-the-shelf data augmentation schemes.
Besides the SimCLR augmentations, we study (i). the $L_p$ corruptions \citep{siedel2023investigating}; (ii). AlgoMix: a mixture of five algorithmic corruptions from CIFAR10-C/CIFAR100-C; and (iii). AugMix \citep{hendrycks2019augmix}. 
With the AlgoMix augmentation, we inject some prior knowledge of the shift between the source and target domain through the augmentations. To be specific, among the 15 corruptions in CIFAR10-C, one out of a set of the first five corruptions is randomly selected and applied to the sample. In this case, we assume partial \textit{a priori} knowledge that the target data are corrupted, but we only know 5 of those corruptions, and we do not know exactly in what way each sample was corrupted. 

Generally, the more similar the augmentation is to the source-target shift, the greater the performance gains observed. On the five datasets with the same corruptions as AlgoMix, the average error drops from 14.70\% using the SimCLR augmentations to 14.00\%, implying that prior knowledge on the shift between the source and target domains would improve accuracy. Across all 15 corruptions, the average error is higher than using SimCLR, but is still better than the baselines. That is, even if the augmentations do not correspond to the true domain shift, removing the nuisances, beats the baselines. 
The AugMix augmentations show the best improvements across the 15 corruptions types, averaging 12.78\% error rate, while the $L_p$ corruptions reaches the lowest error of 20.98\% on the mixed corruption dataset. The results on CIFAR10-C are summarized in Table \ref{tab:aug_scheme}. The last row shows the error when each batch of data is augmented randomly with one of the four schemes. \\

\begin{table}[h]
\centering
\caption{Error  (\%) on CIFAR10-C using different augmentations.}
\label{tab:aug_scheme}
    \begin{tabular}{l c c}
         \toprule
         \multirow{2}{*}{\textbf{Augmentation $\mathcal{A}$}} & \multicolumn{2}{c}{\textbf{CIFAR10-C}}\\
         \cmidrule(lr){2-3}
         & \textbf{Average} & \textbf{Mix 5} \\
         \midrule
         \textbf{SimCLR} \citep{chen2020simple} & 13.13 & 23.68 \\ 
         \textbf{$L_p$ corruption} \citep{siedel2023investigating} & 14.31 & \textbf{20.98} \\
         \textbf{AlgoMix} & 13.92 & 26.25 \\ 
         \textbf{AugMix} \citep{hendrycks2019augmix} & \textbf{12.78} & 25.66 \\
         \textbf{All combined} & 13.10 & 23.13 \\
         \bottomrule
    \end{tabular}
\end{table}

\section{Discussion}

\paragraph{Statistical Significance}

To verify the statistical signifcance of TIRNU's performance, we conducted one-sided Welch’s t-tests \citep{welch1947generalization} comparing our method with SHOT—the strongest baseline in our study.
We selected the six test configurations reported in Table~\ref{tab:corrupt} and repeated SHOT five times to record the mean and standard deviation of its performance. We performed one-sided tests where the null hypothesis assumes equal performance and the alternative assumes one method is better. In Table~\ref{tab:hypo_test} below, we report the results along with the corresponding $p$-values for each configuration. Note that a low $p$-value (below the 5\% significance level) indicates statistical significance.

\begin{table}[h]
    \caption{Comparison of error and statistical significance between SHOT and TIRNU.}
    \centering
    \resizebox{\textwidth}{!}{%
    \begin{tabular}{l|c c c }
         \toprule
         Dataset & SHOT Error ($\pm$ std) & TIRNU Error ($\pm$ std) & $p$-value \\
         \midrule
         CIFAR10 Avg & $15.21 \;(\pm 0.02)$ & $\mathbf{13.13 \;(\pm 0.32)}$ & $6.3\times 10^{-5}$ \\
         CIFAR10 Mix 5 & $25.77 \;(\pm 0.22)$ & $\mathbf{23.68 \;(\pm 0.18)}$ & $1.4\times 10^{-7}$ \\
         CIFAR10 Mix 10 & $26.59 \;(\pm 0.18)$ & $\mathbf{25.69 \;(\pm 0.32)}$ & $6.6\times 10^{-4}$ \\
         CIFAR100 Avg & $38.24 \;(\pm 0.02)$ & $\mathbf{36.68 \;(\pm 0.08)}$ & $2.5\times 10^{-7}$ \\
         CIFAR100 Mix 5 & $52.69 \;(\pm 0.17)$ & $\mathbf{51.62 \;(\pm 0.04)}$ & $4.2\times 10^{-5}$ \\
         CIFAR100 Mix 10 & $\mathbf{98.84 \;(\pm 0.42)}$ & $99.08 \;(\pm 0.10)$ & $0.14$ \\
         \bottomrule
    \end{tabular}
    }
    \label{tab:hypo_test}
\end{table}

As shown, in all settings where TIRNU outperforms SHOT, the p-values are well below the 5\% threshold, confirming the improvements are statistically significant. In the only case where SHOT slightly outperforms TIRNU (CIFAR100 Mix 10), the $p$-value is 0.14, indicating the improvement of SHOT is not statistically significant. These findings support the robustness of TIRNU’s improvements over a strong baseline.

\paragraph{Computational Resources}
TIRNU requires more computational time compared to its competitors, except for MEMO. The extra cost is due to the need for augmentation and mutual information estimation. As we utilize an efficient kernel-based algorithm, the additional computational cost is largely due to the need for augmentation. The computation time for TIRNU per epoch on CIFAR10-C with NVIDIA A40 is 25s, compared to Tent (10s), and SHOT (16s). In the online setting, MEMO requires many augmentations (32 by default) per test sample, and takes about 10 minutes, but ours only need 40s.

\begin{figure}
    \centering
    \includegraphics[width=0.65\textwidth]{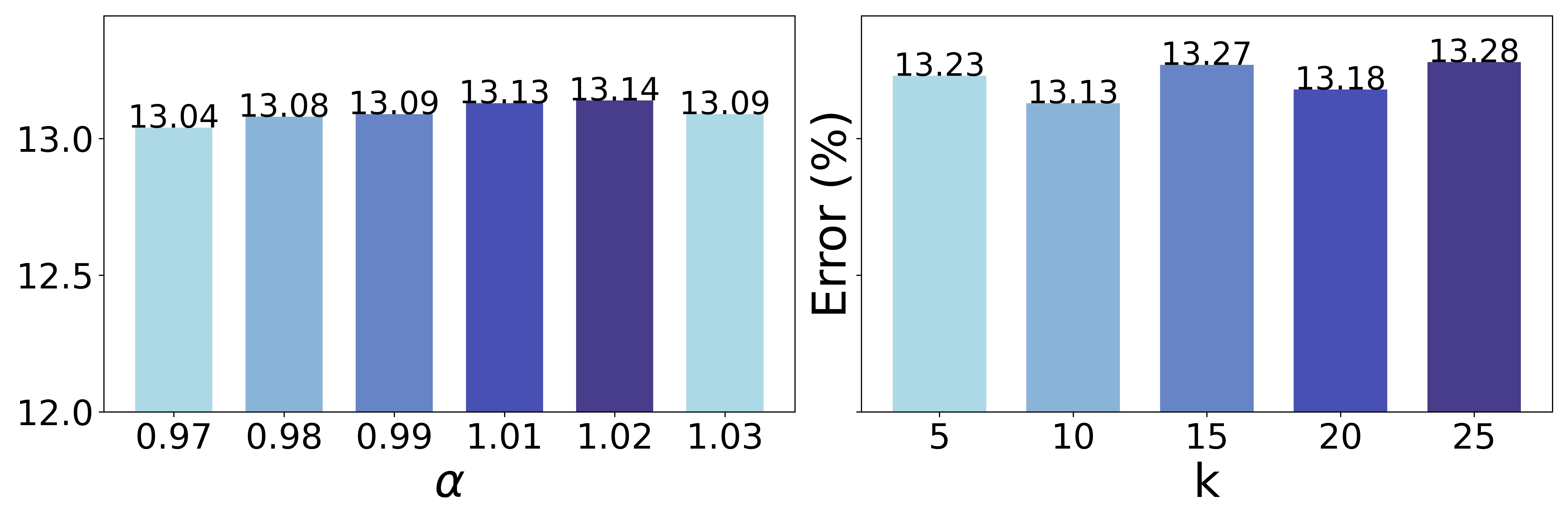}
    \caption{Average error (\%) for the CIFAR10-C dataset for (left) different values of $\alpha$, and (right) different values of $k$. The model performance is not sensitive to either of these two hyperparameters.} 
    \label{fig:meib_par}
\end{figure}

\paragraph{Hyperparameter Finetuning} Our proposed approach requires finetuning of important hyperparameters. For the matrix-based estimation algorithm, $\alpha$ is required. In addition, the Gaussian kernel $\kappa$ is parameterized by the width parameter $\sigma^2$. We compute the average of the $k$ nearest distances from each sample, and choose $\sigma^2$ to be the average over all samples \citep{zhang2022multi}. We empirically testified that our model accuracy is not sensitive to $\alpha$ and $k$ on the CIFAR10-C dataset~(Fig.~\ref{fig:meib_par}). Thus, we set $\alpha=1.01$ and $k=10$ on all other datasets. Nonetheless, for optimal performance, finetuning on specific datasets may be necessary.

For the weight on the overall objective, we keep $\lambda$ fixed at 1, and instead vary the weight on the nuisance unlearning loss in Eqn~\ref{eq:overall_objective} to reach the balance between the two loss terms. In our study, we did not exhaustively fine-tune this weight; instead we found that TIRNU is insensitive to this weight with values between 0.1 and 1.0. Although, when a value of 2 or greater is used, performance degrades, indicating that placing too much weight on unlearning nuisance may misguide the model. Therefore, we adopt the uniform weight of 0.1 in the offline experiments, and 1.0 in the online experiments.

\section{Conclusion}
\label{sec:conclusion}

In this paper, we present Agnostic FTTA, a novel formulation that extends distribution alignment strategies from DA into FTTA, enabling the incorporation of available domain-shift priors for effective model generalization under FTTA constraints. 
Based on this foundation, we introduce TIRNU, a general framework to enhance test-time model adaptability through our innovative ``uncover and unlearn nuisance" strategy.
Emphasizing the modelling of source-target shifts that drive the domain gap, AFTTA reframes the challenging FTTA tasks into a nuisance-shift uncovering and unlearning problem.
Our proposed nuisance unlearning criterion that mitigates nuisance influences in the feature space provides a robust and general solution for improving model performance in test-time adaptation.
Empirically, our TIRNU consistently achieves superior performance across diverse domain adaptation scenarios relative to state-of-the-art methods. 
However, adaptation with TIRNU incurs extra computational costs due to the augmentations and the computation of mutual information, though additional experiments in Section~\ref{subsec:augmentation} indicate that a minimal number of augmentations is sufficient. 
We also note the need to fine-tune the hyperparameter $\lambda$ balancing the loss terms, although this is common practice in deep learning.
Our framework is flexible and can be further adapted with tailored nuisance-shifts that incorporate prior knowledge of source-target domain shifts, thereby enhancing model adaptation with task-specific feature invariance.
For future work, we will explore the potential of such augmentation techniques, such as style transfer, to further boost performance, opening new avenues for research in domain adaptation.

\backmatter

\bmhead{Acknowledgements}

This research is supported by the National Research Foundation, Singapore and DSO National Laboratories under the AI Singapore Programme - “Design Beyond What You Know: MaterialInformed Differential Generative AI (MIDGAI) for LightWeight High-Entropy Alloys and Multi-functional Composites (Stage 1a)” (AISG Award No. AISG2-GC-2023-010).
This research project is supported by the National Research Foundation, Singapore and Infocomm Media Development Authority under its Trust Tech Funding Initiative (No. DTC-RGC-05). Any opinions, findings and conclusions or recommendations expressed in this material are those of the authors and do not reflect the views of National Research Foundation, Singapore and Infocomm Media Development Authority.
This research is supported by A*STAR Career Development Fund <Project No. C243512010>.
P.S. contributed to this publication whilst serving his internship with A*STAR CFAR under the CIARE and Local Student Attachment scheme.

\bibliography{sn-bibliography}%

\setcounter{mainfig}{\value{figure}}
\setcounter{maintab}{\value{table}}

\end{document}